\documentclass{edm_article}
\usepackage{tabularx}
\usepackage{multirow}
\usepackage{rotating, booktabs}
\usepackage{multicol, blindtext}
\usepackage{placeins}
\usepackage{hyperref}
\usepackage{balance}
\usepackage{graphicx}
\usepackage{amsmath}
\usepackage{amssymb}
\usepackage{hyperref}
\usepackage{xcolor}
\usepackage{booktabs}
\usepackage{mathtools}
\usepackage{array}
\usepackage{caption}
\usepackage[export]{adjustbox}
\usepackage{makecell}

\newlength{\imagewidth}
\newlength{\equationwidth}
\usepackage{rotating}
\newcolumntype{P}[1]{>{\centering\arraybackslash}p{#1}}
\newcolumntype{M}[1]{>{\centering\arraybackslash}m{#1}}
\usepackage{multirow}
\usepackage{subfig}
\usepackage{placeins}

\begin{document}
\title{Modeling and Analyzing Scorer Preferences \\ in Short-Answer Math Questions}
\numberofauthors{3} 
\author{
\alignauthor
Mengxue Zhang\\
        \affaddr{UMass Amherst}\\
        \email{mengxuezhang@umass.edu}
\alignauthor Neil Heffernan\\
        \affaddr{Worcester Polytechnic Institute}\\
        \email{nth@wpi.edu}
\and  
\alignauthor Andrew Lan\\
        \affaddr{UMass Amherst}\\
        \email{andrewlan@cs.umass.edu}
}

\maketitle

\begin{abstract}
Automated scoring of student responses to open-ended questions, including short-answer questions, has great potential to scale to a large number of responses. Recent approaches for automated scoring rely on supervised learning, i.e., training classifiers or fine-tuning language models on a small number of responses with human-provided score labels. However, since scoring is a subjective process, these human scores are noisy and can be highly variable, depending on the scorer. In this paper, we investigate a collection of models that account for the individual preferences and tendencies of each human scorer in the automated scoring task. We apply these models to a short-answer math response dataset where each response is scored (often differently) by multiple different human scorers. We conduct quantitative experiments to show that our scorer models lead to improved automated scoring accuracy. We also conduct quantitative experiments and case studies to analyze the individual preferences and tendencies of scorers. We found that scorers can be grouped into several obvious clusters, with each cluster having distinct features, and analyzed them in detail.
\end{abstract}
\section{Introduction}

Automated scoring (AS), i.e., using algorithms to automatically score student (textual) responses to open-ended questions, has significant potential to complement and scale up human scoring, especially with an ever-increasing number of students. AS algorithms are often driven by \emph{supervised} machine learning-based algorithms and require a small number of example responses and their score labels to train on. These algorithms mostly consist of two components: a \emph{representation} component that use either hand-crafted features~\cite{erater,iea,cohmetrix,aes,relevance,semantic} or language models~\cite{cambium,mayfield2020should,taghipour2016neural,uto2020neural,yang2020enhancing} to represent the (mostly textual) content in questions, student responses, and other information, e.g., rubrics~\cite{condor2022representing} and a \emph{scoring} component that use classifiers~\cite{attali2006automated,danielle} to predict the score of a response from its textual representation. In different subject domains, the representation component can be quite different, from hand-crafted features and neural language model-based textual embeddings in automated essay scoring (AES)~\cite{asap,aes}, automatic short answer grading (ASAG)~\cite{irtasag,chiasag}, and reading comprehension scoring~\cite{nigel} to specialized representations in responses where mathematical expressions are present~\cite{baral2022enhancing,mathgpt,mathbertcrap,wang2021scientific}. On the contrary, the scoring model does not vary significantly across different subject domains, often relying on simple classifiers such as logistic regression, support vector machines, random forests, or linear projection heads in neural networks~\cite{dlbook}. We provide a more detailed discussion on related work in Section~\ref{sec:rw}. 

One key factor that limits the accuracy of AS methods is that the scoring task is a \emph{subjective} one; human scorers are often given a set of rubrics~\cite{naepchal} and asked to score responses according to them. However, different individuals interpret rubrics and student responses differently, leading to significant variation in their scores. For example, inter-scorer agreement can be as quite high in NAEP reading comprehension question scoring, with a quadratic weighted Kappa (QWK) score of $0.88$~\cite{nigel} and quite low in open-ended math question scoring, with a Kappa score of $0.083$ (see Section~\ref{sec:data} for details and Table~\ref{tab:example} for a concrete example). This variation creates a \emph{noisy labels} problem, which is a common problem in machine learning where one often needs to acquire a large number of labels via crowdsourcing~\cite{survey-new,survey,github}. In educational applications such as AS, this problem is even more important since the amount of labels we have access to is often small, which amplifies the negative impact of noisy score labels. Therefore, there is a significant need to analyze the preferences and tendencies of individual scorers, to not only improve AS accuracy by providing cleaner labels to train on but also understand where the variation in scores comes from and investigate whether we can reduce it.

\subsection{Contributions}
In this paper, we propose a collection of models for the variation in human scorers due to their individual preferences and tendencies, from simple models that use only a few parameters to account for the bias and variance of each scorer to complex models that use a different set of neural network parameters for each scorer. We ground our work in an AS task for short-answer mathematical questions and show that by adding our model to the classification component of AS models, we can improve AS accuracy by more than $0.02$ in Kappa score and $0.01$ in AUC compared to AS methods that do not account for individual scorer differences. We also conduct qualitative experiments and case studies to analyze the individual preference and tendencies of scorers. We found that scorers can be grouped into several major, obvious clusters, with each cluster having distinct features, which we explain in detail. 
\textbf{We emphasize that our goal is NOT to develop the most accurate AS model; instead, our goal is to show that accounting for the variation across different individual scorers can potentially improve the accuracy of any AS model.}

\subsection{Related work}
\label{sec:rw}

\paragraph{Noisy labels} 
Individual scorers often exhibit different preferences and tendencies, as found in \cite{wang2017latent}. Some of our models for scorer preference and tendency are closely related to models used in peer grading~\cite{joachims}, where students grade each others' work, which is often deployed in settings such as massive open online courses (MOOCs) where a large number of open-ended responses make it impossible for external human scorers to score all responses. Most of these models are inspired by methods in machine learning on combining labels from human labelers with different expertise in crowdsourcing contexts~\cite{whitehill2009whose}. These models are simple and interpretable, with the most basic version involving a single bias parameter (towards certain score labels) and a single variance parameter (across different score labels) for each scorer. On the contrary, we experiment with not only these models but also more flexible but uninterpretable models, which are compatible with using pre-trained neural language models~\cite{devlin2018bert,radford2019gpt2} in the representation component of AS models. 

\paragraph{AS and math AS} 
The majority of existing ASAG and AES methods focus on non-mathematical domains \cite{basu2013powergrading,chen2013automated,pardos,cohmetrix,aes,semantic,wang2019meta}. Recently, some AS methods are developed for specific domains that contain non-textual symbols, e.g., Chemistry, Computer Science, and Physics, which exist in student responses in addition to text, achieving higher and higher AS accuracy \cite{sami,erickson,leacock2003c,srikant2014system,taghipour2016neural}. 
Our work is grounded in the short-answer math question scoring setting, which is studied in prior works~\cite{sami,baral2022enhancing,mathbertcrap,edm22}. The key technical challenge here is that mathematical expressions that are often contained in open-ended student responses can be difficult to parse and understand in the representation component. The authors of \cite{sami} proposed a scoring approach for short-answer math questions using sentence-BERT (SBERT)-based representation of student responses and simply ignored mathematical expressions. The authors of \cite{baral2022enhancing} developed an additional set of features specifically designed for mathematical expressions and used them in conjunction with the SBERT representations as input to the scoring component. The authors of \cite{mathbertcrap} fine-tuned a language model, BERT \cite{devlin2018bert}, further pre-trained on math textbooks, as the representation component; however, this representation was found to not be highly effective in later works \cite{edm22}. The authors of \cite{edm22} used a sophisticated in-context meta-training approach for automated scoring by inputting not only the response that needs to be scored but also scored examples to a language model, enabling the language model to learn from examples, which results in significant improvement in AS accuracy and especially generalizability to previously unseen questions. 

Another line of related work is about fairness in educational data analysis since scorer preference can be classified as a form of individual bias. Researchers have proposed methods to incorporate constraints and regularization into predictive models to improve parity and mitigate fairness issues \cite{chu2022mitigating,Zafar:Fairness:2017,Zemel:Learning:2013}. On the contrary, our work does not attempt at reducing biases; our focus is only on identifying a specific source of bias, individual scorer bias, in the AS context. Therefore, the only approach we use to mitigate biases is to leverage scorer identification information and investigate its impact on AS accuracy, following prior work on using this information in predictive models \cite{yu2021should}. 

\section{Model}

We now detail our models for individual scorer preference and tendency in AS tasks. For all models, we use a BERT model~\cite{devlin2018bert} as the corresponding representation component of the AS model, which has been shown to perform well and reach state-of-the-art performance on the short math answer AS task with an appropriate input structure~\cite{edm22}. Let us denote each question-response pair that needs to be scored as $q_i$, while the $j$-th scorer assigns a score $y_{i,j} \in \{1, \ldots, C\}$ where $C$ denotes the number of possible score categories. 

\subsection{Baseline}

Our base AS model is one that directly uses the output \texttt{[CLS]} embedding of BERT as the representation of the question-response pair $\mathbf{r}_i \in \mathbb{R}^D$, where $D=768$ is the dimension of the embedding. We also use a linear classification head (omitting the bias terms for simplicity) with softmax output~\cite{dlbook} for all score categories, i.e.,
    \begin{align*}
        p(y_{i,j} = c) \propto e^{(\mathbf{w}_c^T \mathbf{r}_i)+b_c},
    \end{align*}
where $\mathbf{w}_c$ denotes the $D$-dimensional parameter for each score category and $b_c \in \mathbb{R}$ is the universal bias toward each score category.

\begin{table*}[h]
\caption{Example questions, student responses, and scores. Some scorers assign highly different scores to similar responses.}
\resizebox{2\columnwidth}{!}{
\begin{tabularx}{\textwidth}{ l|X|X|l|l }
  \toprule
  $question\_id$ & $question\_body$  & $response$  & $scorer\_id$ & $score$ \\ \hline
  43737 & Chris spent \$9 of the \$12 he was given for his birthday. His sister Jessie says that he has spent exactly 0.75 of the money. Chris wonders if Jessie is correct. Explain your reasoning. & 
  Jessie is correct because 0.75 in fraction form is 3/4. 9 is 3/4 of 12, so she is right. & 1 &4 \\
  \cline{3-5}
  && Jessie is wrong. & 1 & 0\\
\cline{3-5}
  && she is correct & 1&1 \\
  \cline{3-5}
  && Jessie is incorrect.& 2 & 4 \\ 
  \cline{3-5} 
  && Jessie is right because if you divide 12 by 9 you get 0.75. & 2 & 2\\
  \bottomrule
\end{tabularx}}
\vspace{.2cm}

\label{tab:example}
\end{table*}

\subsection{Scalar bias and variance with scorer embeddings}
\label{sec:semb}
The first version of our model is the simplest and most interpretable: we use a scalar temperature, i.e., variance parameter for each scorer, and a scalar offset, i.e., bias parameter on each score category for each scorer, i.e.,
\begin{align} \label{eq:scalar}
        p(y_{i,j} = c) \propto e^{\alpha_j (\mathbf{w}_c^T \mathbf{r}_i + b_{c,j})},
\end{align}
where $\alpha_t > 0$ is the ``temperature'' parameter that controls the scorer's uncertainty across categories: larger values indicate higher concentrations of the probability mass around the most likely score category, which corresponds to more consistent scoring behavior. $b_{c,j} \in \mathbb{R}$ is the ``offset'' parameter that controls the scorer's bias towards each score category: larger values indicate a higher probability of selecting some score category, which corresponds to more positive/negative scoring preferences. 

In practice, we found that parameterizing biases with a set of \emph{scorer embeddings} lead to better performance than simply parameterizing the biases as learnable scalars. Specifically, we introduce a high-dimensional embedding for each scorer, $\mathbf{e}_j \in \mathbb{R}^D$, and use a $C \times D$ matrix $\mathbf{S}$ to map it to a low-dimensional vector that corresponds to the bias terms for all score categories. This advantage is likely due to the fact that more model parameters make the model more flexible and more capable in capturing detailed nuances in scorer preferences and tendencies. 

\subsection{Content-driven scorer bias and variance}

\sloppy
In the models above, we have set the scorer biases and variances to be scorer-dependent but not question/response-dependent, i.e., the bias and variance of a scorer stay the same across all question-response pairs. However, in practice, it is possible that these parameters depend on the actual textual content of the question and the student's response. Therefore, we extend the scorer model in Eq.~\ref{eq:scalar} into
\begin{align*}
    & \mathbf{b}_{i,j} = f_b(\mathbf{r}_i,\mathbf{e}_j), \quad 
    \alpha_t = f_{\alpha}(\mathbf{r}_i,\mathbf{e}_j), \quad \\ &
    \text{where} \quad  f_b(\mathbf{r}_i,\mathbf{e}_j) = \mathbf{r}_i^T \mathbf{A}_b \mathbf{e}_j, \quad
    f_{\alpha}(\mathbf{r}_i,\mathbf{e}_j) = \mathbf{r}_i^T \mathbf{A}_{\alpha} \mathbf{e}_j, 
\end{align*}
where the bias $\mathbf{b}_{i,j}$ is now a $C \times 1$ vector of biases across all score categories and both question-response pair ($i$)-dependent and scorer ($j$)-dependent. $f_b$ and $f_\alpha$ denote functions that map the textual representation of the question-response pair and the scorer embedding to the bias and variance parameters, which can be implemented in any way (from simple linear models to complex neural networks). In this work, we found that using bi-linear functions of the question-response pair representation $\mathbf{r}_i$ and the scorer embedding $\mathbf{e}_j$, using two $D \times D$ matrices $\mathbf{A}_b$ and $\mathbf{A}_\alpha$, results in the best AS accuracy. 


\subsection{Training with different losses}
We explore using various different loss functions as objectives to train our AS model, which we detail below. 

\subsubsection{Cross-entropy}
Since the AS task corresponds to a multi-category classification problem, the standard loss function that we minimize is the cross-entropy (CE) loss~\cite{dlbook}, summed over all question-response pairs and scorers, as
\begin{align*}
    \mathcal{L}_\text{CE} = -\sum_{i,j} \sum_{c=1}^C \mathbf{1}_{y_{i,j} = c} \log p(y_{i,j} = c)
\end{align*}
where $\mathbf{1}_{y_{i,j} = c}$ is the indicator function that is non-zero only if $y_{i,j}=c$. In other words, we are minimizing the negative log-likelihood of the actual score category among the category probabilities predicted by the AS model, $p(y_{i,j} = c)$. 

\subsubsection{Ordinal log loss}
One obvious limitation of the standard CE loss is that it assumes that the categories are unordered, which works for many applications. Therefore, it penalizes all misclassifications equally. However, for AS, the score categories are naturally ordered, which means that score classification errors are not equal: if the actual score is $1$ out of $5$, then a misclassified score of $2$ is better than $5$, but they are weighted equally in the standard CE loss. Therefore, we follow the approach outlined in \cite{ordlog} and use an ordinal log loss (OLL), which we define as
\begin{align*}
    \mathcal{L}_\text{OLL} = -\sum_{i,j} \sum_{c=1}^C |y_{i,j}-c| \log (1 - p(y_{i,j}=c)),
\end{align*}
where we weight the misclassification likelihood, i.e., $-\log (1 - p(y_{i,j}=c))$, according to the difference between the actual score, $y_{i,j}$, and the predicted score, $c$. In the aforementioned example, this objective function would increase the penalty of a misclassified score of $5$ by four times compared to a misclassified score of $2$ when the actual score is $1$, which effectively leverages the ordered nature of the score categories.

\subsubsection{Mean squared error}

Since the score categories are integers and can be treated as numerical values, one simple alternative to the CE loss is the mean squared error (MSE) loss, i.e., 
\begin{align} \label{eq:mse}
  \mathcal{L}_\text{MSE} = \sum_{i,j} (y_{i,j} - \sum_{c=1}^C p(y_{i,j}=c)c)^2,
\end{align}
where we simply square the difference between the actual score and the expected (i.e., weighted average) score under the category probabilities predicted by the AS model.

\begin{table*}[]
\caption{Comparing different scorer models on short-answer math scoring. The combination of content-driven scorer bias and temperature with the OLL loss outperforms other scorer models and training losses.}
\resizebox{2\columnwidth}{!}{
\begin{tabular}{l|l|l|l|l}
\hline
Bias ($b$) \& Temperature ($\alpha$) & Loss Function & AUC & RMSE & Kappa \\ \hline
Universal ($b_c$, $\alpha$ = 1)  & CE            & 0.765 $\pm$ 0.003    &0.954 $\pm$ 0.014      & 0.614 $\pm$ 0.009      \\ \hline
Universal ($b_c$, $\alpha$ = 1)  & MSE            & 0.764 $\pm$ 0.003    &0.946 $\pm$ 0.018      & 0.615 $\pm$ 0.008      \\ \hline
Universal ($b_c$, $\alpha$ = 1)  & OLL            & 0.768 $\pm$ 0.003    &0.944 $\pm$ 0.015      & 0.617 $\pm$ 0.006      \\ \hline
Scorer-specific ($b_{c,j}$, $\alpha_j$)      & CE            & 0.768 $\pm$ 0.005    &   0.928 $\pm$ 0.023   &   0.628 $\pm$ 0.006    \\ \hline
Scorer-specific  ($b_{c,j}$, $\alpha_j$)        & MSE    &  0.772 $\pm$ 0.005   &  0.926 $\pm$ 0.025    &  0.625  $\pm$  0.006  \\ \hline
Scorer-specific   ($b_{c,j}$, $\alpha_j$)       & OLL           &  0.770 $\pm$ 0.003   &   \textbf{0.916 $\pm$ 0.013}   &  0.628 $\pm$ 0.004     \\ \hline
Content-driven ($b_{c,j}(\mathbf{r}_i)$, $\alpha_j(\mathbf{r}_i)$)        & CE            &    0.772 $\pm$ 0.003  & 0.923 $\pm$ 0.016     &  0.631 $\pm$ 0.006      \\ \hline
Content-driven ($b_{c,j}(\mathbf{r}_i)$, $\alpha_j(\mathbf{r}_i)$)        & MSE            &    0.774 $\pm$ 0.004 &   0.922 $\pm$ 0.021   & 0.629 $\pm$ 0.005      \\ \hline
Content-driven ($b_{c,j}(\mathbf{r}_i)$, $\alpha_j(\mathbf{r}_i)$)       & OLL           &   \textbf{0.779 $\pm$ 0.004}  &  0.924 $\pm$ 0.013    &    \textbf{0.641 $\pm$ 0.005}   \\ \hline
\end{tabular}}
\vspace{.2cm}
\label{tab:as}
\end{table*}

\section{Quantitative Experiments}

We now detail experiments that we conducted to validate the different scoring components of AS models and loss functions that capture scorer preferences and tendencies. Section~\ref{sec:data} discusses details on the real-world student response dataset we use and the pre-processing steps. Section~\ref{metrics} details the evaluation metrics we use in our experiments. Section~\ref{setting} details our experimental setting, and Section~\ref{result} details the experimental results and corresponding discussion. 

\subsection{Dataset}
\label{sec:data}

We use data collected from an online learning platform that has been used in prior work~\cite{sami,erickson}, which contains student responses to open-ended, short-answer math questions, together with scores assigned by human scores. 
There are a total of 141,612 total student responses made by $25,069$ students to $2,042$ questions, with $891$ different teachers being scorers. The set of possible score categories is from $0$ (no credit) to $4$ (full credit). The dataset mainly contains math word problems, where the answer could be mathematical such as numbers and equations or textual explanations, sometimes in the format of images. 

We found that different scorers sometimes assign very different scores to the same response, which motivated this work. As an example, we analyze question-response pairs that are scored by more than one scorer and evaluate the Kappa score between these scorers. The \emph{human} Kappa score is only $0.083$, which means a minimal agreement between different scorers. Although there are only $523$ such pairs, this case study still shows that even for the same exact response, scorers have highly different individual preferences and tendencies and may assign them highly different scores. 

We also perform a series of pre-processing steps to the original dataset. For example, since some of the scorers do not score many responses, e.g., less than $100$, there may not be enough information on these scorers for us to model their behavior. Therefore, we remove these scores from the dataset, which results in $203$ scorers, $1,273$ questions, and $118,079$ responses. The average score is $3.152 \pm 1.417$. Table~\ref{tab:example} shows some examples of data points of this dataset; each data point consists of the question statement, the student's response, the scorer's ID, and the score.

\subsection{Metrics}
\label{metrics}
We utilize three standard evaluation metrics for integer-valued scores that have been commonly used in the AS task \cite{sami,erickson}. First, the area under the receiver operating characteristic curve (\textbf{AUC}) metric, which we adapt to the multi-category classification problem by averaging the AUC numbers over each possible score category and treating them as separate binary classification problems, following \cite{hand2001simple}. Second, we use the root mean squared error (\textbf{RMSE}) metric, which simply treats the integer-valued score categories as numbers, as detailed in Eq.~\ref{eq:mse}. Third and most importantly, we use the multi-class Cohen's \textbf{Kappa} metric for ordered categories, which is often used to evaluate AS methods \cite{naepchal}. 

\begin{figure*}%
    \centering
    \subfloat[\centering 2-D visualization of the learned scorer embedding space ]{{\includegraphics[width=0.5\textwidth]{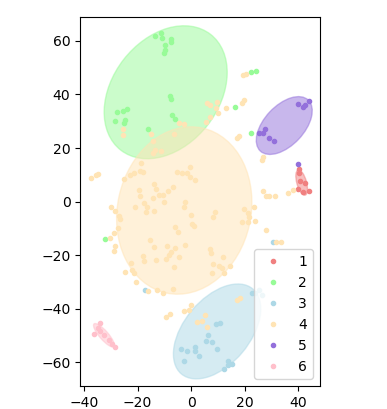}}}%
    \hspace{-.5cm}
    \subfloat[\centering Bias for each score category]{{\includegraphics[width=0.5\textwidth]{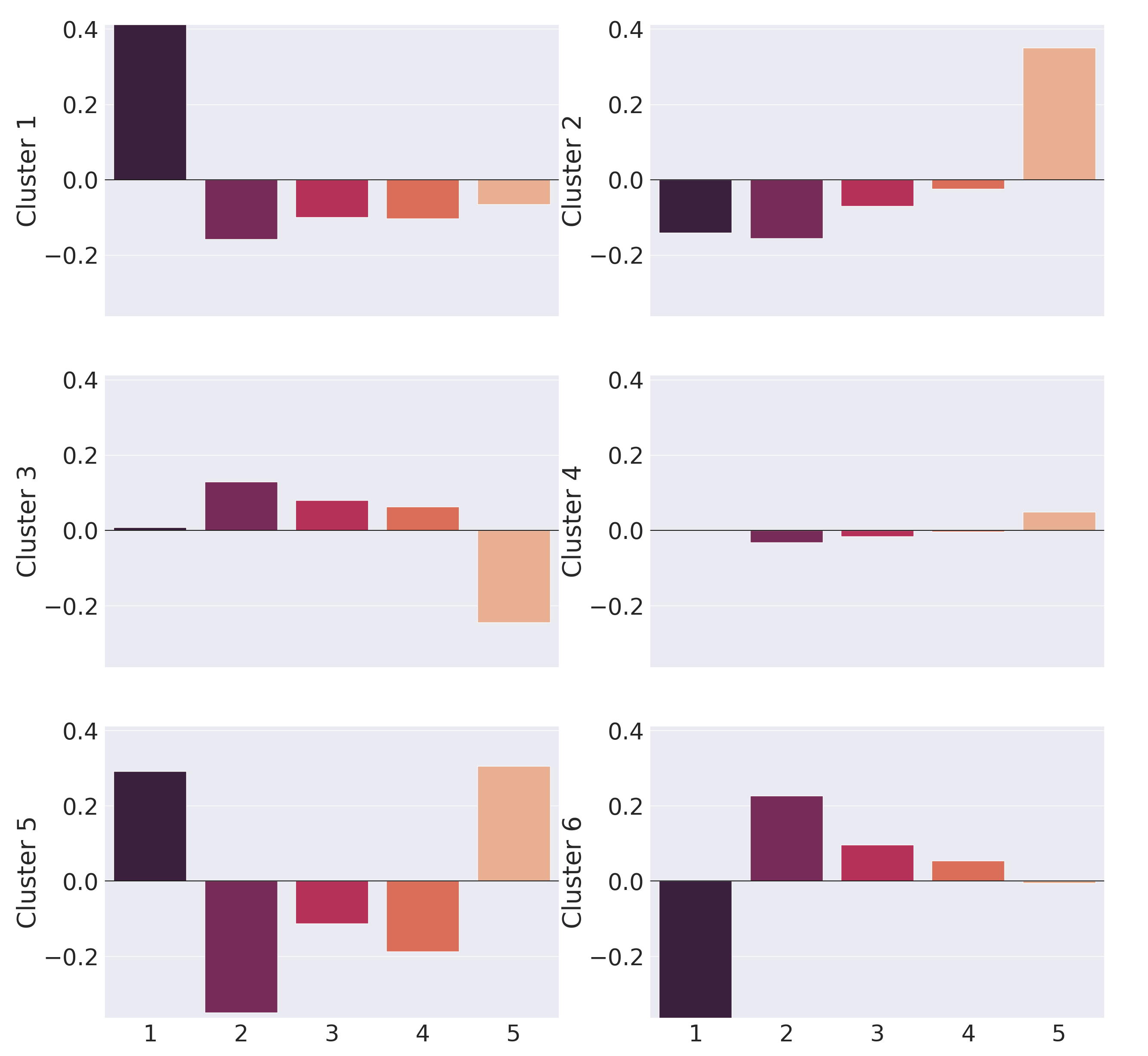}}}%
    \caption{Visualization of clustering result on scorer embedding learnt via scorer-specific model. The left figure shows the 2-D visualization of scorer embedding space, and the right figure shows the average bias for each cluster}%
    \label{fig:emb_bcj}%
\end{figure*}

\subsection{Experimental setting}
\label{setting}

In the quantitative experiment, we focus on studying whether adding scorer information leads to improved AS accuracy. Therefore, when we are splitting a dataset into training, validation, and test sets, we ensure that every scorer is included in the training set. We divide the data points (question-response pairs, scorer ID, score) into $10$ equally-sized folds for cross-validation. 
During training, we use $8$ folds as training data, $1$ fold for validation for model selection, and $1$ fold for the final testing to evaluate the AS models. 

For a fair comparison, every model uses BERT\footnote{\url{https://huggingface.co/bert-base-uncased}} as the pre-trained model for question-response pair representation, which has been shown to result in state-of-the-art AS accuracy in prior work~\cite{edm22}. We emphasize that our work on \textbf{scorer models} can be added on top of \textbf{any} AS method for response representation; applying these models on other AS methods is left for future work. We use the Adam optimizer, a batch size of $16$, and a learning rate of $1e-5$ for $10$ training epochs on an NVIDIA RTX$8000$ GPU. We do not perform any hyper-parameter tuning and simply use the default settings.

\subsection{Results and discussion}
\label{result}

Table~\ref{tab:as} shows the mean and standard deviation of each scorer model trained under each loss function. We see that generally, models with content-driven scorer biases and variances outperform scorer-specific biases and variances, which outperform the base AS model that treats each scorer the same with universal values for bias and variance. The improvement in AS accuracy is significant, up to about $0.02$ in the most important metric--Kappa, for the content-driven biases and variances over the standard AS approach of not using scorer information. This observation validates the need to account for individual scorer preferences and tendencies in the highly subjective AS task. Meanwhile, since the content-driven scorer bias and variance models outperform the scorer-specific bias and variance models, we can conclude that the content of the question and response does play an important role in scorer preference. 

We also observe that training scorer models with the OLL loss outperform the other losse, while training with the MSE loss does not even lead to the best results on the RMSE metric. This observation suggests that taking into account the ordered nature of score categories instead of treating them as parallel ones is important to the AS task.

\section{Qualitative Analysis} 

Despite the content-driven model delivering the highest AUC and Kappa results, the complexity of the information contained in its embedding space renders it difficult to interpret. Consequently, we have elected to concentrate on examining the scorer-specific model (detailed in Sec.~\ref{sec:semb}).  

\subsection{Visualization of scorer embedding}
Figure~\ref{fig:emb_bcj} shows a 2-D visualization of the learned scorer embedding space; We see that there are obvious clusters among all scorers. We then fit the learned scorer embeddings under a mixture-of-Gaussian model via the expectation-maximization (EM) algorithm with $6$ clusters. The subfigures to each side of the main plot shows each cluster's average bias towards each score category, which are $0$, $1$, $2$, $3$, and $4$ from left to right. 


\begin{table*}[ht]  
  \caption{Detailed biases and variance (inverse of temperature) for each scorer profile, their observed scoring distributions, and average response features. We normalize the observed scoring distributions to zero-mean, which makes them easier to visually compare against the learned biases. \textit{math tok (\%)} is the percentage of math tokens in the response. \textit{img (\%)} is the percentage of images in the response. \textit{length} is the number of word tokens in the response.}
\resizebox{2\columnwidth}{!}{
  \begin{tabular}
      {M{0.08\textwidth}|
      M{0.15\textwidth} 
      M{0.15\textwidth} 
      M{0.1\textwidth} 
      M{0.1\textwidth} %
      M{0.1\textwidth}
      M{0.1\textwidth}
      M{0.1\textwidth}}
      
      \hline 
      Cluster & 
      Bias & 
      Observed scoring distribution (normalized) & 
      Temperature & 
      Score  & 
      \multicolumn{3}{c}{Response features}
      \\\cline{6-8}
       &&&&&math tok (\%)&img (\%)&length\\
      
      \hline
      1
      & \parbox[l]{0.15\textwidth}{\includegraphics[width=0.15\textwidth]{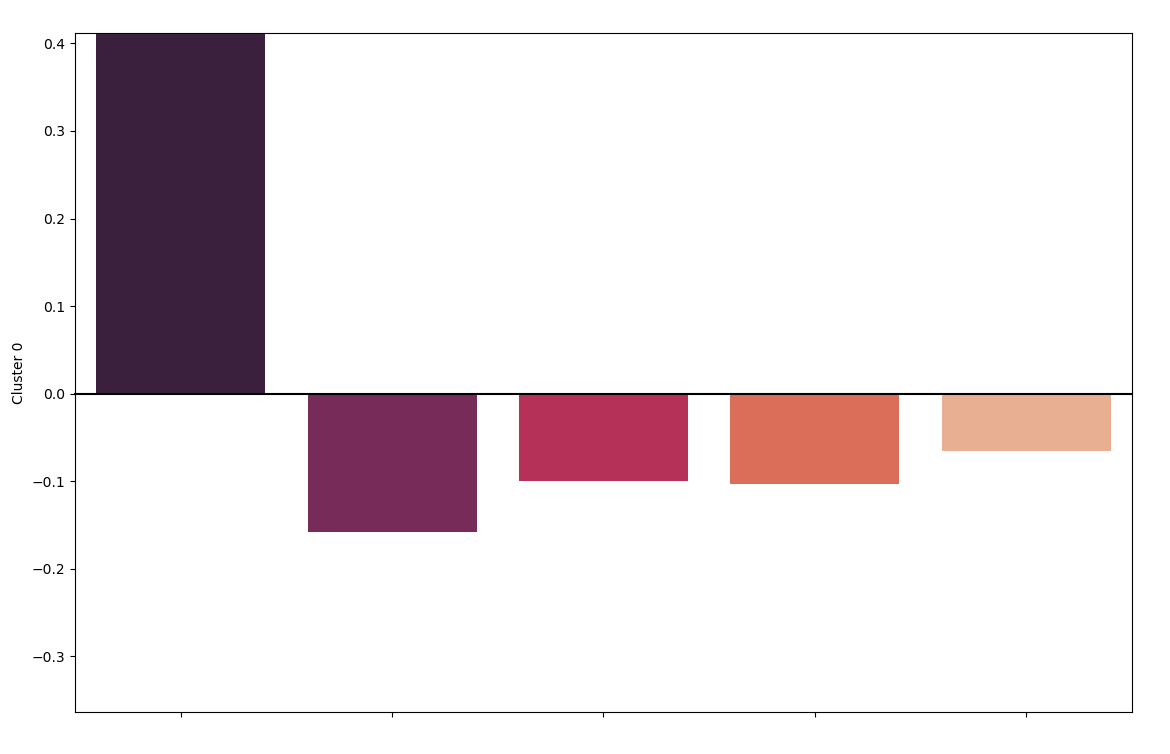}} 
      &\parbox[l]{0.13\textwidth}{\includegraphics[width=0.13\textwidth]{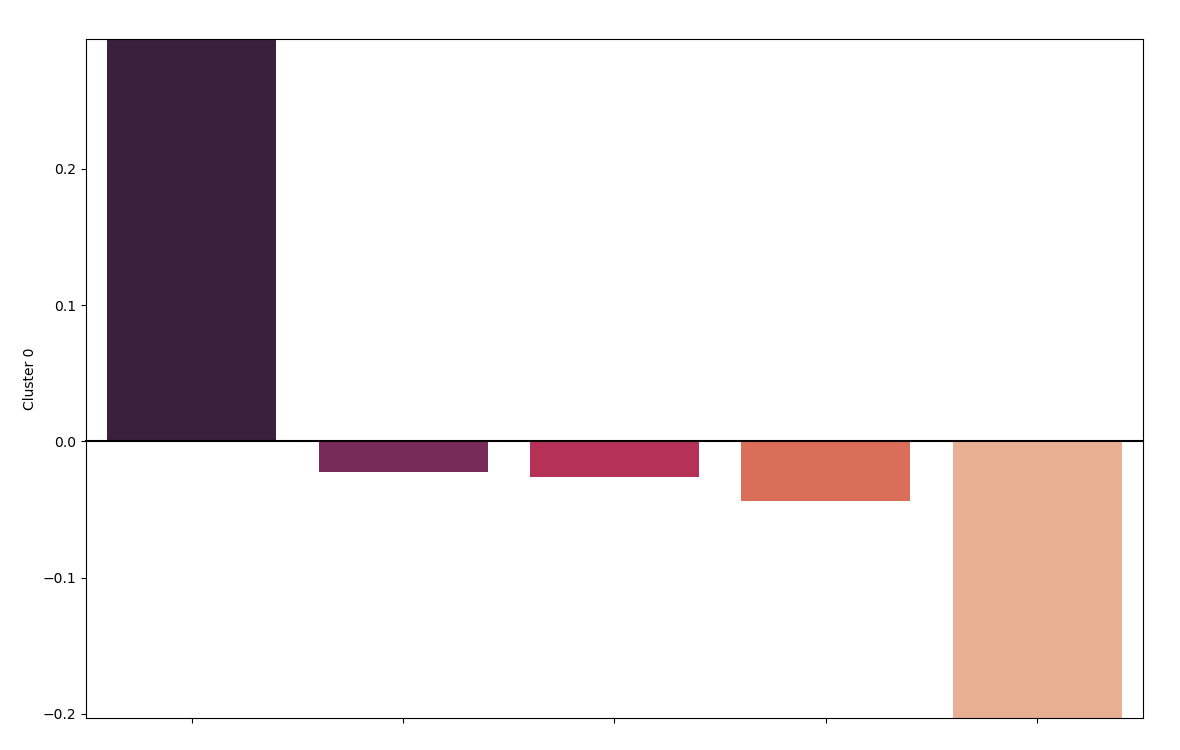}} 
      & 1.013
      &  1.685 $\pm$ 1.644 
      & 29.13 
      & 0.101 
      & 23.06
      \\
      \hline
      2
      & \parbox[l]{0.15\textwidth}{\includegraphics[width=0.15\textwidth]{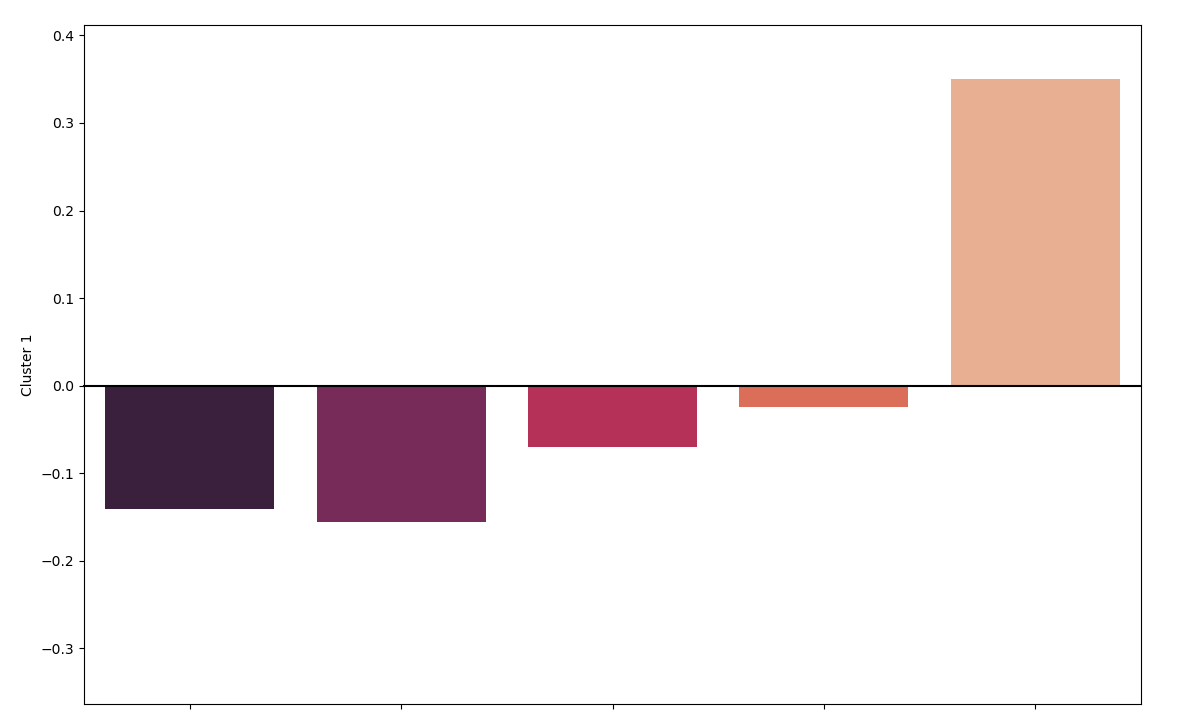}} 
      &\parbox[l]{0.15\textwidth}{\includegraphics[width=0.15\textwidth]{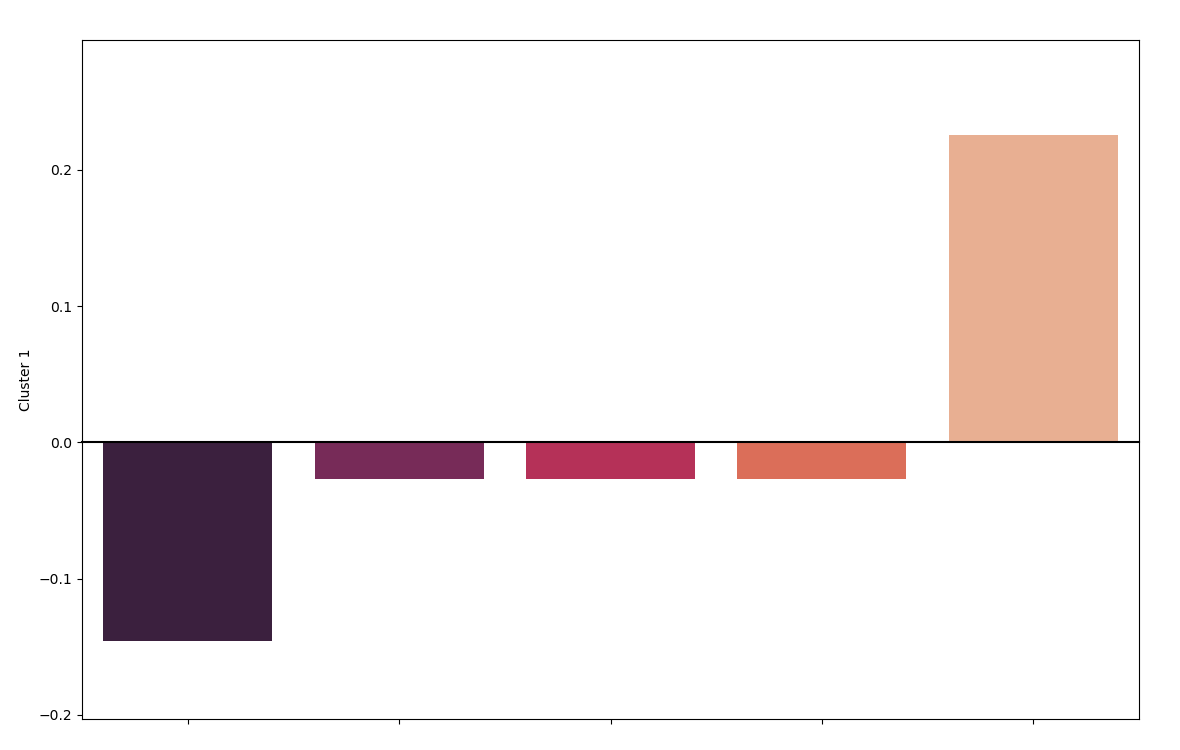}} 
      & 1.034
      &  3.451 $\pm$ 0.919
      & 32.12 
      & 1.286
      & 24.40
      \\
      \hline
      3
      & \parbox[l]{0.15\textwidth}{\includegraphics[width=0.15\textwidth]{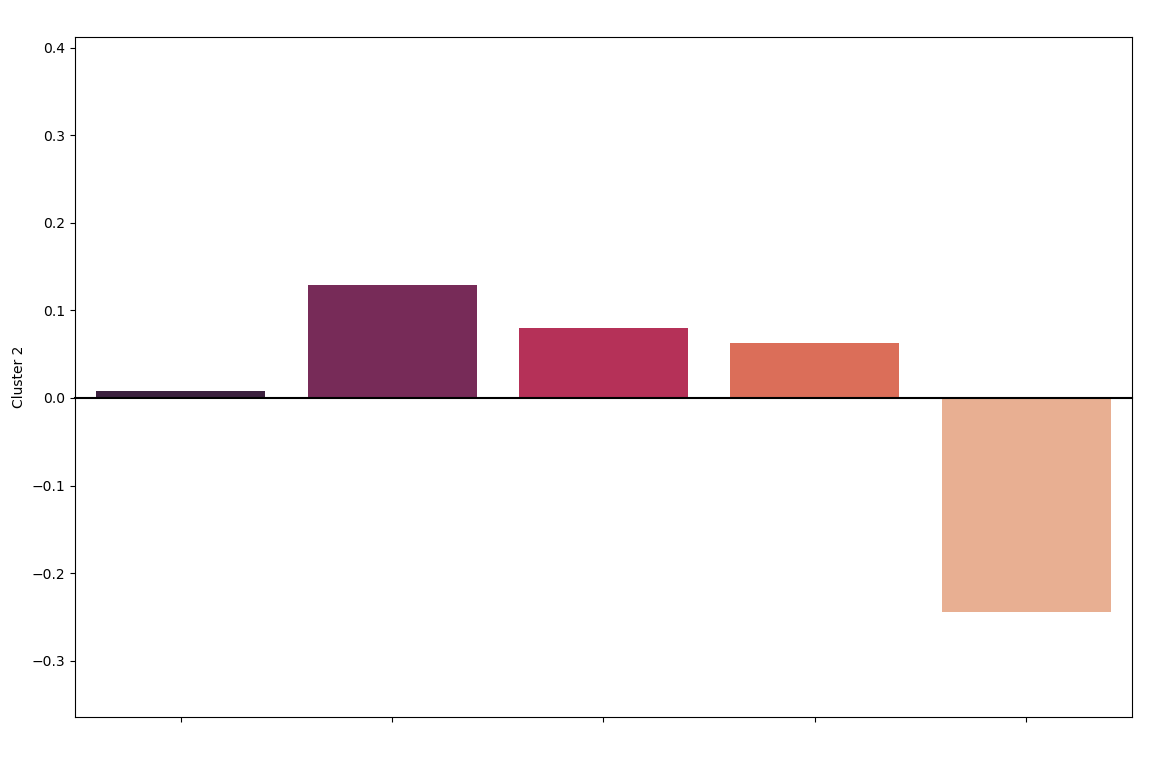}} 
      &\parbox[l]{0.15\textwidth}{\includegraphics[width=0.15\textwidth]{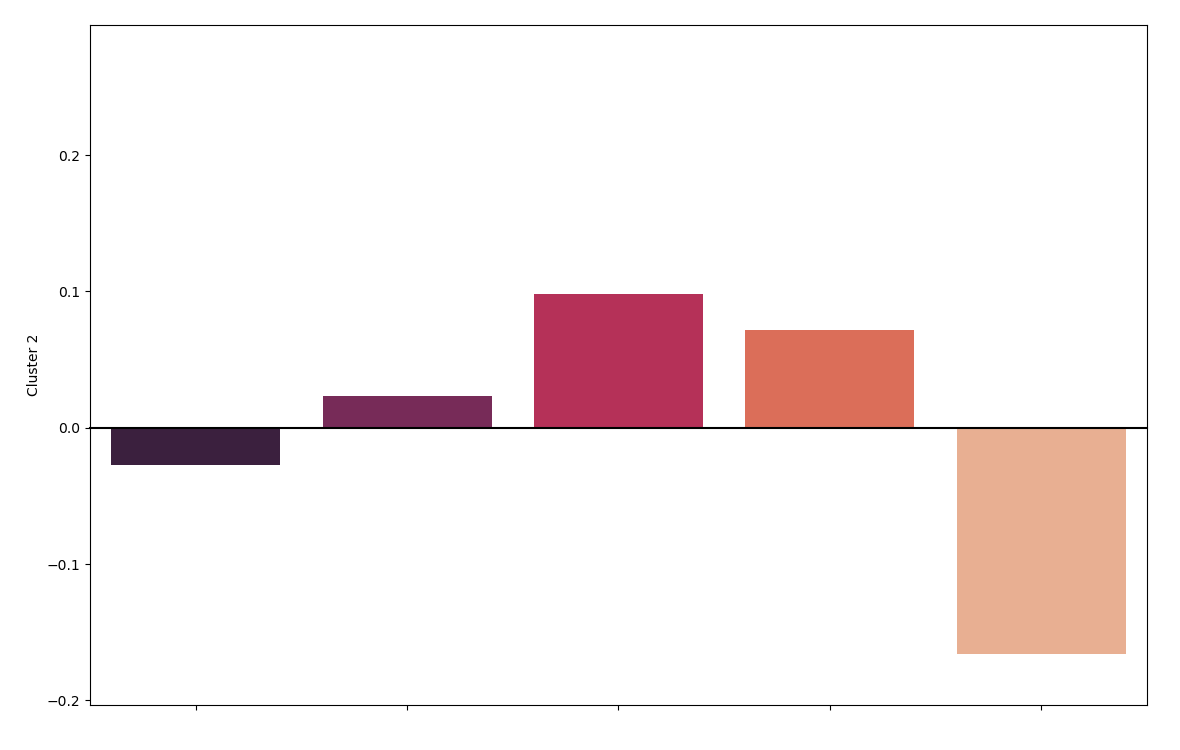}} 
      & 0.996
      &  2.415 $\pm$ 1.400 
      & 23.51 
      & 1.311
      & 36.16
      \\
      \hline
            4
      & \parbox[l]{0.15\textwidth}{\includegraphics[width=0.15\textwidth]{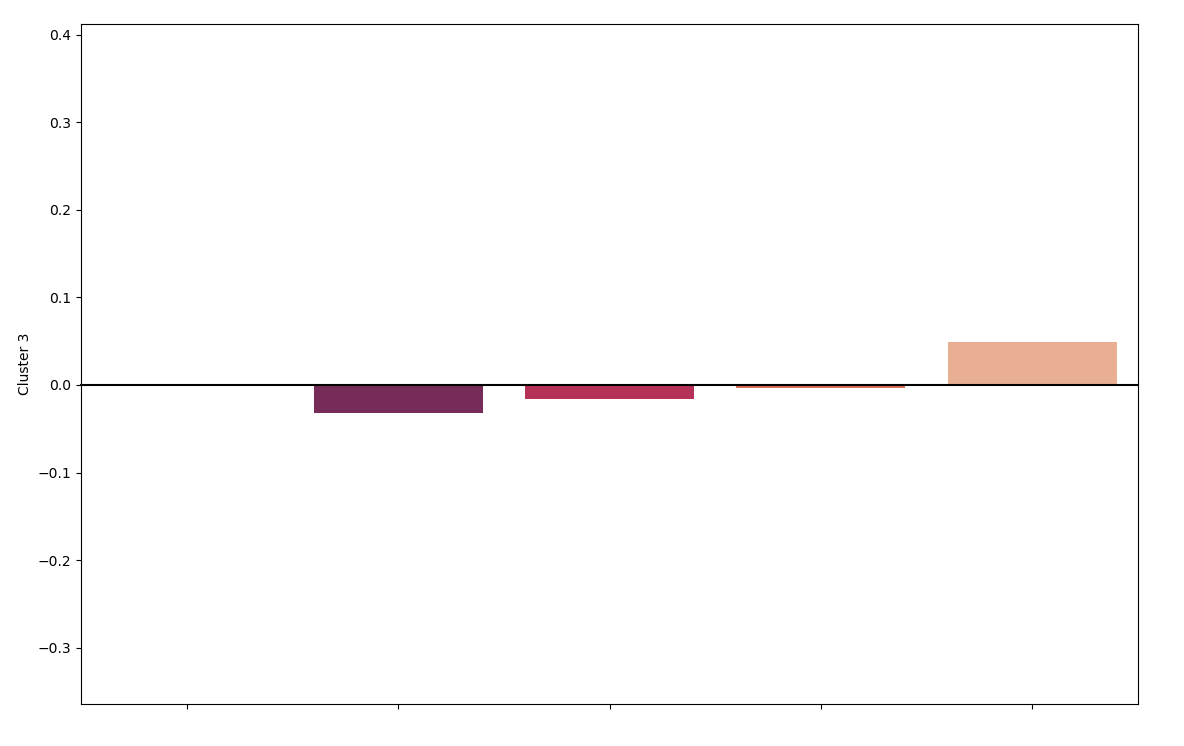}} 
      &\parbox[l]{0.15\textwidth}{\includegraphics[width=0.15\textwidth]{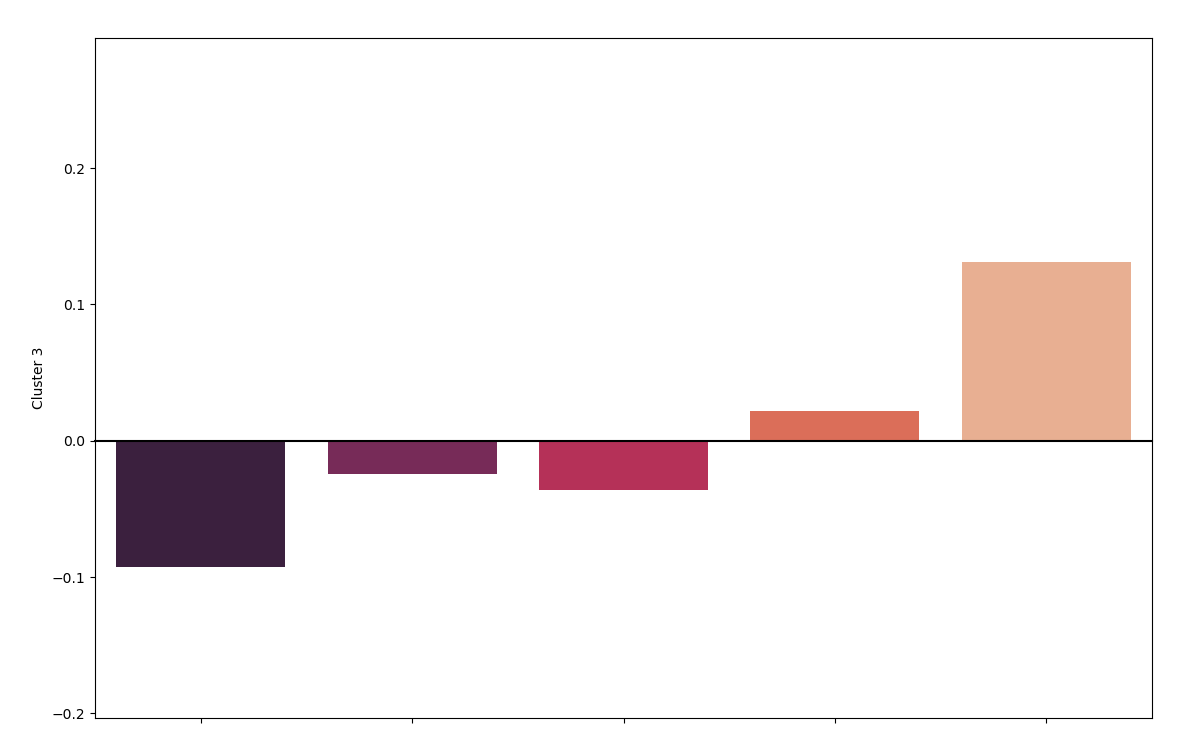}} 
      & 1.033
      &  3.074 $\pm$ 0.991 
      & 29.48 
      & 0.304
      & 21.94
      \\
      \hline
            5
      & \parbox[l]{0.15\textwidth}{\includegraphics[width=0.15\textwidth]{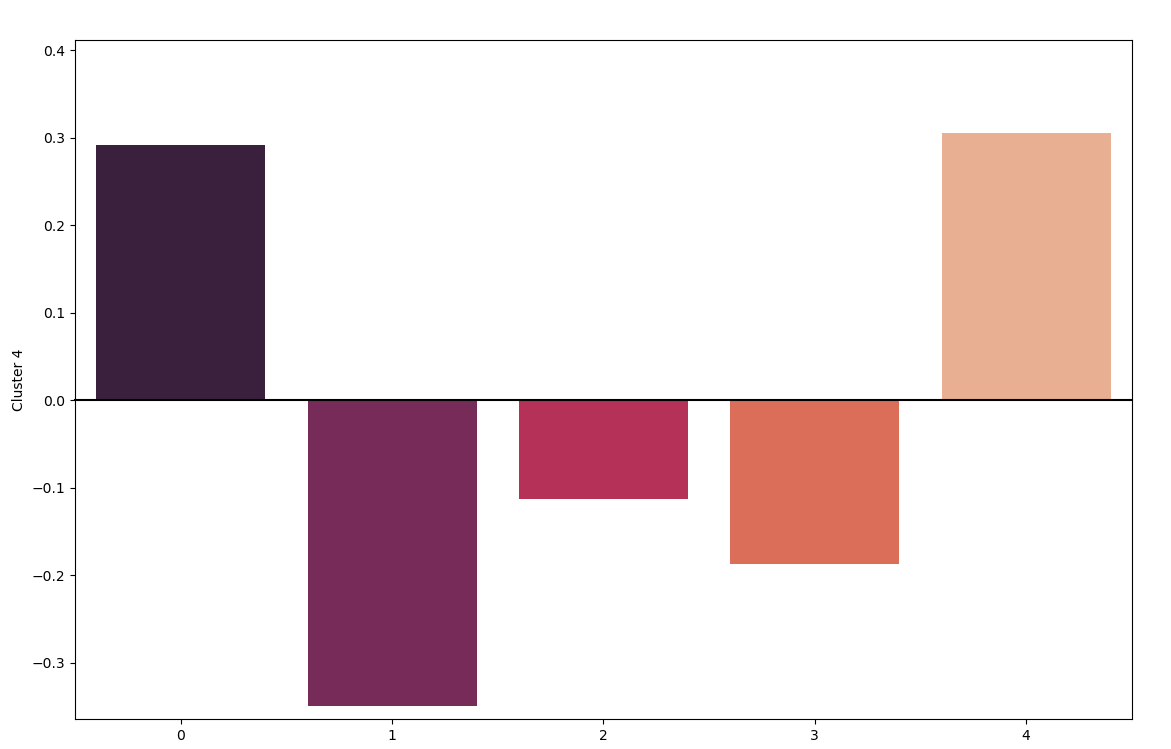}} 
      &\parbox[l]{0.15\textwidth}{\includegraphics[width=0.15\textwidth]{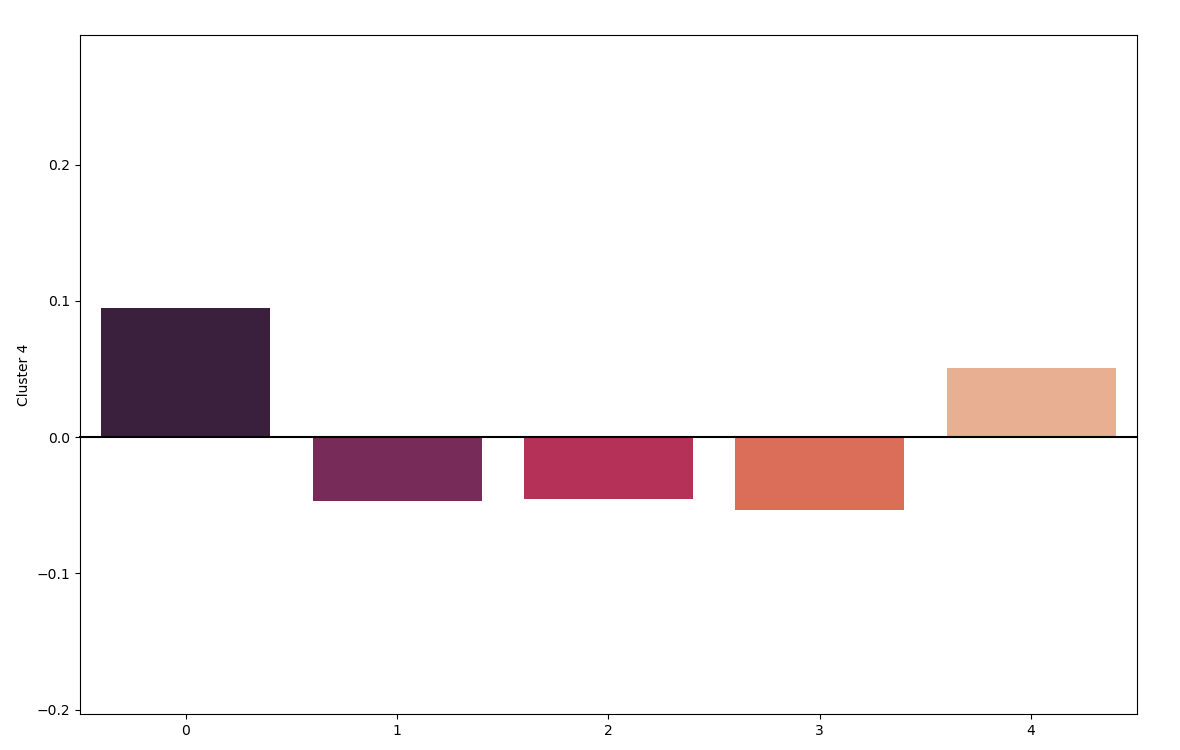}} 
      & 1.026
      &  2.558 $\pm$ 1.806
      & 45.18
      & 5.271
      & 14.35
      \\
      \hline
      6
      & \parbox[l]{0.15\textwidth}{\includegraphics[width=0.15\textwidth]{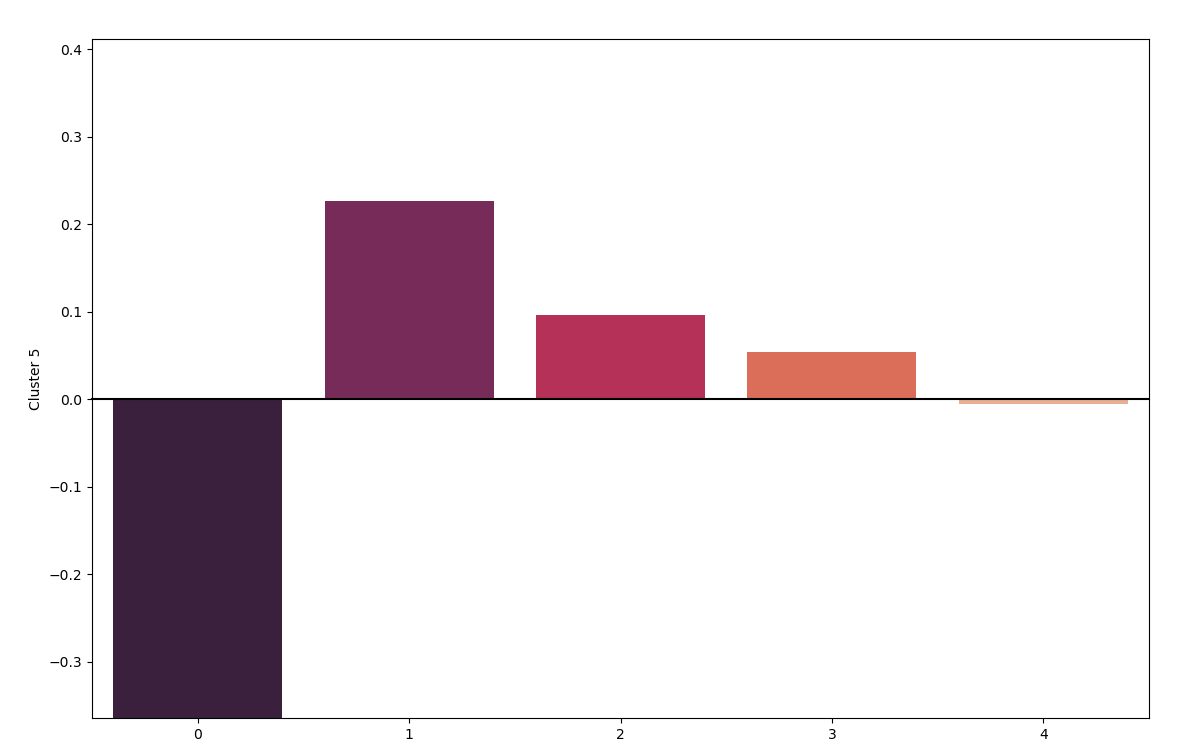}} 
      &\parbox[l]{0.15\textwidth}{\includegraphics[width=0.15\textwidth]{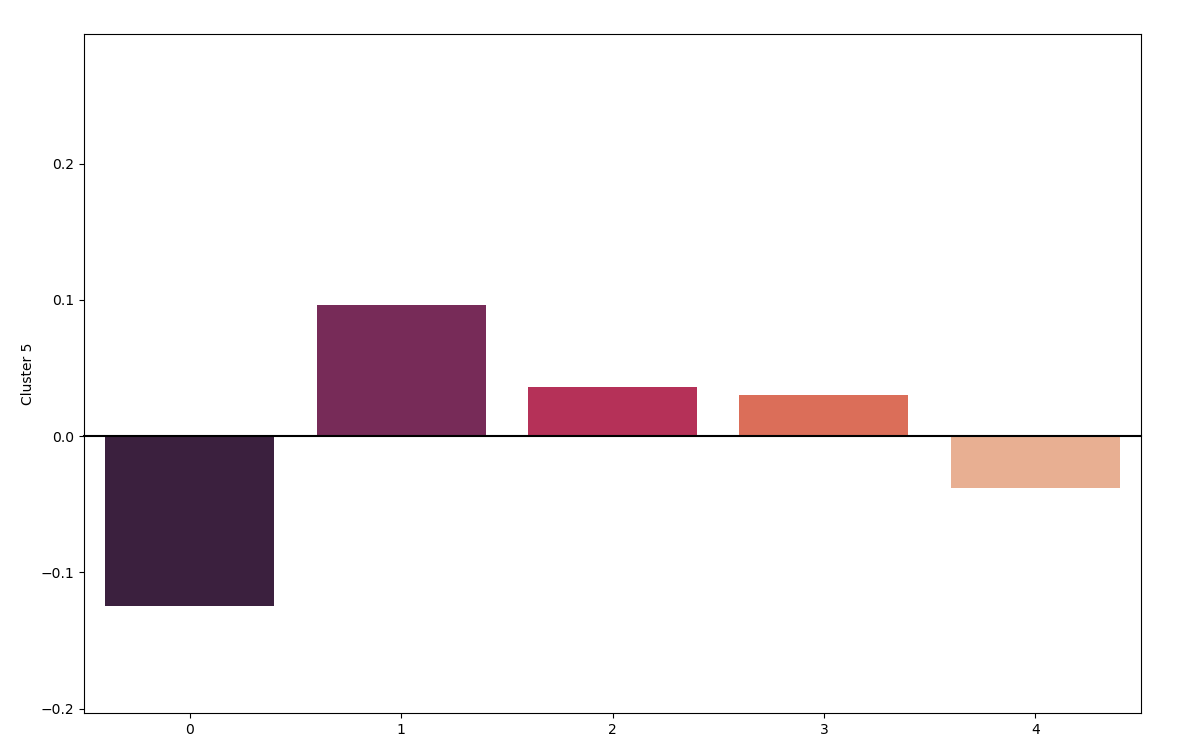}} 
      & 1.007
      &  2.714 $\pm$ 1.331
      & 33.83 
      & 1.403
      & 13.34
      \\
      \hline
  \end{tabular}}
  \vspace{-.2cm}
  \label{tab:bcj}
\end{table*}

\subsection{Features analysis based on each cluster}

Cluster $1$ shows a negative scoring profile, with a strong, positive bias towards the lowest score category $0$ (positive $b_{c,j}$ values) and small, negative biases against higher scores, $1$, $2$, and $3$ (negative $b_{c,j}$ values). These scorers assign $0$ scores much more often than other score categories, compared to other scorers. The average score across question-response pairs is the lowest for this cluster, at $1.69$. Meanwhile, this cluster has a relatively high score variance of $1.69$, meaning that these scorers tend to have inconsistent behavior and assign a wide variety of score labels.  

Cluster $2$ shows a positive scoring profile, with a strong, positive bias towards the highest score, $4$, and moderate negative biases against other scores. These scorers prefer to assign scores that are overwhelmingly higher compared to other scorers. The average score across question-response pairs is the lowest for this cluster, at $3.45$. Meanwhile, this cluster has a relatively low score variance of $0.92$, meaning that these scorers are consistent in scoring responses higher than other scorers.  

Cluster $3$ shows a conservative scoring profile, with small, positive biases towards the middling scores $1$, $2$, and $3$ and a strong, negative bias against the top score $4$. The average score across question-response pairs is $2.41$ for this cluster with a variance of $1.4$, which is high considering that scorers in this cluster rarely use the top score category, indicating that their scoring behavior is not highly consistent. 

Cluster $4$ shows an unbiased scoring profile, with a low bias towards or against any score category, with a slight preference for the top score category, $4$. This cluster contains almost half of the scorers, which means that the majority of scorers are reliable (their scores depend mostly on the actual quality of the response, i.e., the $\mathbf{w}_c^T \mathbf{r}_i$ term in Eq.~\ref{eq:scalar} rather than the bias term. 

Cluster $5$ shows a polarizing scoring profile, with strong, positive biases toward both the lowest score, $0$, and the highest score, $4$, while having strong, negative biases against score categories in between. Scorers in this cluster often score a response as all or nothing while using the intermediate score values sparingly. The average score across question-response pairs is $2.55$ for this cluster with a variance of $1.81$, the highest among all clusters, which agrees with our observation that these scorers are highly polarizing and rarely judge any response to be partially correct. 

Cluster $6$ shows a lenient scoring profile, with a strong, negative bias against the lowest score, $0$, and a moderate, positive bias towards the next score, $1$, with minimal bias across higher score categories. Scorers in this cluster tend to award students a single point for an incorrect response instead of no points at all. The average score across question-response pairs is $2.71$ for this cluster with a middling variance of $1.33$.

\sloppy
\section{Conclusions and Future Work}


In this paper, We created models to account for individual scorer preferences and tendencies in short-answer math response automated scoring. Our models differ from previous work by focusing on capturing the subjective nature of scoring rather than textual content. Our models range from simple to complex, with some using bias and variance as a function of the question and response. Our experiments on a dataset with low inter-rater agreement showed that accounting for scorer preferences and tendencies improved performance by more than $0.02$ in the Kappa metric. Qualitative analysis showed obvious patterns among scorers, some with biases towards certain scores. 
Scorer-specific settings can model scorer grading behavior very well. In other words, the scorer's grading behavior is highly controllable, and the scorer's grading behavior representation is also well-represented in the hidden space. One practical extension could be adjusting the learned scorer bias by using a different type of scorer embedding to control model grading in a different scorer style.  
Future work can address limitations in our analysis. Our dataset only provides scorer IDs, lacking gender, race, or location. Investigating biases with this additional information is crucial, including how teacher-student relationships or shared demographics impact biases. Our analysis also did not consider student demographic information, which is important for fairness studies. Additionally, our scorer models were only validated with a BERT-based textual representation model, so further testing is needed to determine their adaptability to traditional, feature-based automated scoring methods.

\section{Acknowledgements}
The authors thank the NSF (under grants 1917713, 2118706, 2202506, 2215193) for partially supporting this work.

\balance
\clearpage
\bibliographystyle{abbrv}
\bibliography{references}

\begin{thebibliography}{10}

\bibitem{naepchal}
The ed.gov national assessment of educational progress (naep) automated scoring
  challenge.
\newblock Online: \url{https://github.com/NAEP-AS-Challenge/info}, 2021.

\bibitem{asap}
The hewlett foundation: Automated essay scoring.
\newblock Online: \url{https://www.kaggle.com/c/asap-aes}, 2021.

\bibitem{survey-new}
G.~Algan and I.~Ulusoy.
\newblock Image classification with deep learning in the presence of noisy
  labels: A survey.
\newblock {\em arXiv preprint arXiv:1912.05170}, 2019.

\bibitem{attali2006automated}
Y.~Attali and J.~Burstein.
\newblock Automated essay scoring with e-rater{\textregistered} v. 2.
\newblock {\em The Journal of Technology, Learning and Assessment}, 4:3, 2006.

\bibitem{sami}
S.~Baral, A.~F. Botelho, J.~A. Erickson, P.~Benachamardi, and N.~T. Heffernan.
\newblock Improving automated scoring of student open responses in mathematics.
\newblock {\em International Educational Data Mining Society}, 2021.

\bibitem{baral2022enhancing}
S.~Baral, K.~Seetharaman, A.~F. Botelho, A.~Wang, G.~Heineman, and N.~T.
  Heffernan.
\newblock Enhancing auto-scoring of student open responses in the presence of
  mathematical terms and expressions.
\newblock In {\em International Conference on Artificial Intelligence in
  Education}, pages 685--690. Springer, 2022.

\bibitem{basu2013powergrading}
S.~Basu, C.~Jacobs, and L.~Vanderwende.
\newblock Powergrading: a clustering approach to amplify human effort for short
  answer grading.
\newblock {\em Transactions of the Association for Computational Linguistics},
  1:391--402, 2013.

\bibitem{erater}
J.~Burstein.
\newblock The e-rater{\textregistered} scoring engine: Automated essay scoring
  with natural language processing.
\newblock 2003.

\bibitem{chen2013automated}
H.~Chen and B.~He.
\newblock Automated essay scoring by maximizing human-machine agreement.
\newblock In {\em Proceedings of the 2013 Conference on Empirical Methods in
  Natural Language Processing}, pages 1741--1752, 2013.

\bibitem{chu2022mitigating}
Y.-W. Chu, S.~Hosseinalipour, E.~Tenorio, L.~Cruz, K.~Douglas, A.~Lan, and
  C.~Brinton.
\newblock Mitigating biases in student performance prediction via
  attention-based personalized federated learning.
\newblock In {\em Proceedings of the 31st ACM International Conference on
  Information \& Knowledge Management}, pages 3033--3042, 2022.

\bibitem{pardos}
A.~Condor, M.~Litster, and Z.~Pardos.
\newblock Automatic short answer grading with sbert on out-of-sample questions.
\newblock {\em International Educational Data Mining Society}, 2021.

\bibitem{condor2022representing}
A.~Condor, Z.~Pardos, and M.~Linn.
\newblock Representing scoring rubrics as graphs for automatic short answer
  grading.
\newblock In {\em International Conference on Artificial Intelligence in
  Education}, pages 354--365. Springer, 2022.

\bibitem{devlin2018bert}
J.~Devlin, M.-W. Chang, K.~Lee, and K.~Toutanova.
\newblock Bert: Pre-training of deep bidirectional transformers for language
  understanding.
\newblock {\em arXiv preprint arXiv:1810.04805}, 2018.

\bibitem{erickson}
J.~A. Erickson, A.~F. Botelho, S.~McAteer, A.~Varatharaj, and N.~T. Heffernan.
\newblock The automated grading of student open responses in mathematics.
\newblock In {\em Proceedings of the International Conference on Learning
  Analytics \& Knowledge}, page 615–624, 2020.

\bibitem{ordlog}
F.~C. et~al.
\newblock A simple log-based loss function for ordinal text classification.
\newblock online:\url{https://openreview.net/pdf?id=khB9is39GvL}, 2022.

\bibitem{nigel}
N.~Fernandez, A.~Ghosh, N.~Liu, Z.~Wang, B.~Choffin, R.~G. Baraniuk, and A.~S.
  Lan.
\newblock Automated scoring for reading comprehension via in-context bert
  tuning.
\newblock In {\em International Conference on Artificial Intelligence in
  Education}, page~0, 2022.

\bibitem{iea}
P.~W. Foltz, D.~Laham, and T.~K. Landauer.
\newblock The intelligent essay assessor: Applications to educational
  technology.
\newblock {\em Interactive Multimedia Electronic Journal of Computer-Enhanced
  Learning}, 1(2):939--944, 1999.

\bibitem{survey}
B.~Fr{\'e}nay and M.~Verleysen.
\newblock Classification in the presence of label noise: a survey.
\newblock {\em IEEE transactions on neural networks and learning systems},
  25(5):845--869, 2013.

\bibitem{github}
Github.
\newblock Awesome-learning-with-label-noise.
\newblock
  \href{https://github.com/subeeshvasu/Awesome-Learning-with-Label-Noise}{https://github.com/subeeshvasu/Awesome-Learning-with-Label-Noise},
  2020.

\bibitem{dlbook}
I.~Goodfellow, Y.~Bengio, and A.~Courville.
\newblock {\em Deep Learning}.
\newblock MIT Press, 2016.

\bibitem{cohmetrix}
A.~C. Graesser, D.~S. McNamara, M.~M. Louwerse, and Z.~Cai.
\newblock Coh-metrix: Analysis of text on cohesion and language.
\newblock {\em Behavior research methods, instruments, \& computers},
  36(2):193--202, 2004.

\bibitem{hand2001simple}
D.~J. Hand and R.~J. Till.
\newblock A simple generalisation of the area under the roc curve for multiple
  class classification problems.
\newblock {\em Machine learning}, 45(2):171--186, 2001.

\bibitem{leacock2003c}
C.~Leacock and M.~Chodorow.
\newblock C-rater: Automated scoring of short-answer questions.
\newblock {\em Computers and the Humanities}, 37(4):389--405, 2003.

\bibitem{cambium}
S.~Lottridge, B.~Godek, A.~Jafari, and M.~Patel.
\newblock Comparing the robustness of deep learning and classical automated
  scoring approaches to gaming strategies.
\newblock Technical report, Cambium Assessment Inc., 2021.

\bibitem{mayfield2020should}
E.~Mayfield and A.~W. Black.
\newblock Should you fine-tune bert for automated essay scoring?
\newblock In {\em 15th Workshop on Innovative Use of NLP for Building
  Educational Applications}, pages 151--162, 2020.

\bibitem{danielle}
D.~S. McNamara, S.~A. Crossley, R.~D. Roscoe, L.~K. Allen, and J.~Dai.
\newblock A hierarchical classification approach to automated essay scoring.
\newblock {\em Assessing Writing}, 23:35--59, 2015.

\bibitem{aes}
E.~B. Page.
\newblock The imminence of grading essays by computer.
\newblock {\em The Phi Delta Kappan}, 47(5):238--243, 1966.

\bibitem{relevance}
I.~Persing and V.~Ng.
\newblock Modeling prompt adherence in student essays.
\newblock In {\em 52nd Annual Meeting of the Association for Computational
  Linguistics}, pages 1534--1543, 2014.

\bibitem{radford2019gpt2}
A.~Radford, J.~Wu, R.~Child, D.~Luan, D.~Amodei, and I.~Sutskever.
\newblock Language models are unsupervised multitask learners.
\newblock {\em OpenAI blog}, 1(8):9, 2019.

\bibitem{joachims}
K.~Raman and T.~Joachims.
\newblock Methods for ordinal peer grading.
\newblock In {\em Proc. 20th ACM SIGKDD Intl. Conf. on Knowledge Discovery and
  Data Mining}, pages 1037--1046, Aug. 2014.

\bibitem{mathgpt}
A.~Scarlatos and A.~Lan.
\newblock Tree-based representation and generation of natural and mathematical
  language.
\newblock In {\em Proceedings of the Annual Meeting of the Association for
  Computational Linguistics (ACL)}, 2023, preprint:
  \url{https://arxiv.org/abs/2302.07974}.

\bibitem{mathbertcrap}
J.~T. Shen, M.~Yamashita, E.~Prihar, N.~Heffernan, X.~Wu, B.~Graff, and D.~Lee.
\newblock Mathbert: A pre-trained language model for general nlp tasks in
  mathematics education.
\newblock {\em arXiv preprint arXiv:2106.07340}, 2021.

\bibitem{srikant2014system}
S.~Srikant and V.~Aggarwal.
\newblock A system to grade computer programming skills using machine learning.
\newblock In {\em Proceedings of the 20th ACM SIGKDD international conference
  on Knowledge discovery and data mining}, pages 1887--1896, 2014.

\bibitem{taghipour2016neural}
K.~Taghipour and H.~T. Ng.
\newblock A neural approach to automated essay scoring.
\newblock In {\em Empirical methods in natural language processing}, pages
  1882--1891, 2016.

\bibitem{irtasag}
M.~Uto and Y.~Uchida.
\newblock Automated short-answer grading using deep neural networks and item
  response theory.
\newblock In {\em International Conference on Artificial Intelligence in
  Education}, pages 334--339, 2020.

\bibitem{uto2020neural}
M.~Uto, Y.~Xie, and M.~Ueno.
\newblock Neural automated essay scoring incorporating handcrafted features.
\newblock In {\em 28th Conference on Computational Linguistics}, pages
  6077--6088, 2020.

\bibitem{semantic}
S.~Valenti, F.~Neri, and A.~Cucchiarelli.
\newblock An overview of current research on automated essay grading.
\newblock {\em Journal of Information Technology Education: Research},
  2(1):319--330, 2003.

\bibitem{wang2017latent}
J.~Z. Wang, A.~S. Lan, P.~J. Grimaldi, and R.~G. Baraniuk.
\newblock A latent factor model for instructor content preference analysis.
\newblock {\em International Educational Data Mining Society}, 2017.

\bibitem{wang2019meta}
Z.~Wang, A.~Lan, A.~Waters, P.~Grimaldi, and R.~Baraniuk.
\newblock A meta-learning augmented bidirectional transformer model for
  automatic short answer grading.
\newblock In {\em Proc. 12th Int. Conf. Educ. Data Mining (EDM)}, pages 1--4,
  2019.

\bibitem{wang2021scientific}
Z.~Wang, M.~Zhang, R.~G. Baraniuk, and A.~S. Lan.
\newblock Scientific formula retrieval via tree embeddings.
\newblock In {\em 2021 IEEE International Conference on Big Data (Big Data)},
  pages 1493--1503. IEEE, 2021.

\bibitem{whitehill2009whose}
J.~Whitehill, T.-f. Wu, J.~Bergsma, J.~Movellan, and P.~Ruvolo.
\newblock Whose vote should count more: Optimal integration of labels from
  labelers of unknown expertise.
\newblock {\em Advances in neural information processing systems}, 22, 2009.

\bibitem{yang2020enhancing}
R.~Yang, J.~Cao, Z.~Wen, Y.~Wu, and X.~He.
\newblock Enhancing automated essay scoring performance via fine-tuning
  pre-trained language models with combination of regression and ranking.
\newblock {\em Findings of the Association for Computational Linguistics:
  EMNLP}, 2020:1560--1569, 2020.

\bibitem{yu2021should}
R.~Yu, H.~Lee, and R.~F. Kizilcec.
\newblock Should college dropout prediction models include protected
  attributes?
\newblock In {\em 8th ACM Conference on Learning@ Scale}, pages 91--100, 2021.

\bibitem{Zafar:Fairness:2017}
M.~B. Zafar, I.~Valera, M.~Gomez~Rodriguez, and K.~P. Gummadi.
\newblock Fairness beyond disparate treatment \& disparate impact: Learning
  classification without disparate mistreatment.
\newblock In {\em International Conference on World Wide Web}, pages
  1171--1180, 2017.

\bibitem{Zemel:Learning:2013}
R.~Zemel, Y.~Wu, K.~Swersky, T.~Pitassi, and C.~Dwork.
\newblock Learning fair representations.
\newblock In {\em International Conference on Machine Learning}, pages
  325--333, 2013.

\bibitem{edm22}
M.~Zhang, S.~Baral, N.~Heffernan, and A.~Lan.
\newblock Automatic short math answer grading via in-context meta-learning.
\newblock {\em arXiv preprint arXiv:2205.15219}, 2022.

\bibitem{chiasag}
Y.~Zhang, R.~Shah, and M.~Chi.
\newblock Deep learning+ student modeling+ clustering: A recipe for effective
  automatic short answer grading.
\newblock In {\em International Conference on Educational Data Mining}, page
  562, 2016.

\end{thebibliography}
\clearpage

\appendix
\section{Correlation analysis}

\begin{figure*}[t]%
    \centering
    \subfloat[\centering Correlation Coefficient Matrix]{{\includegraphics[width=0.5\textwidth]{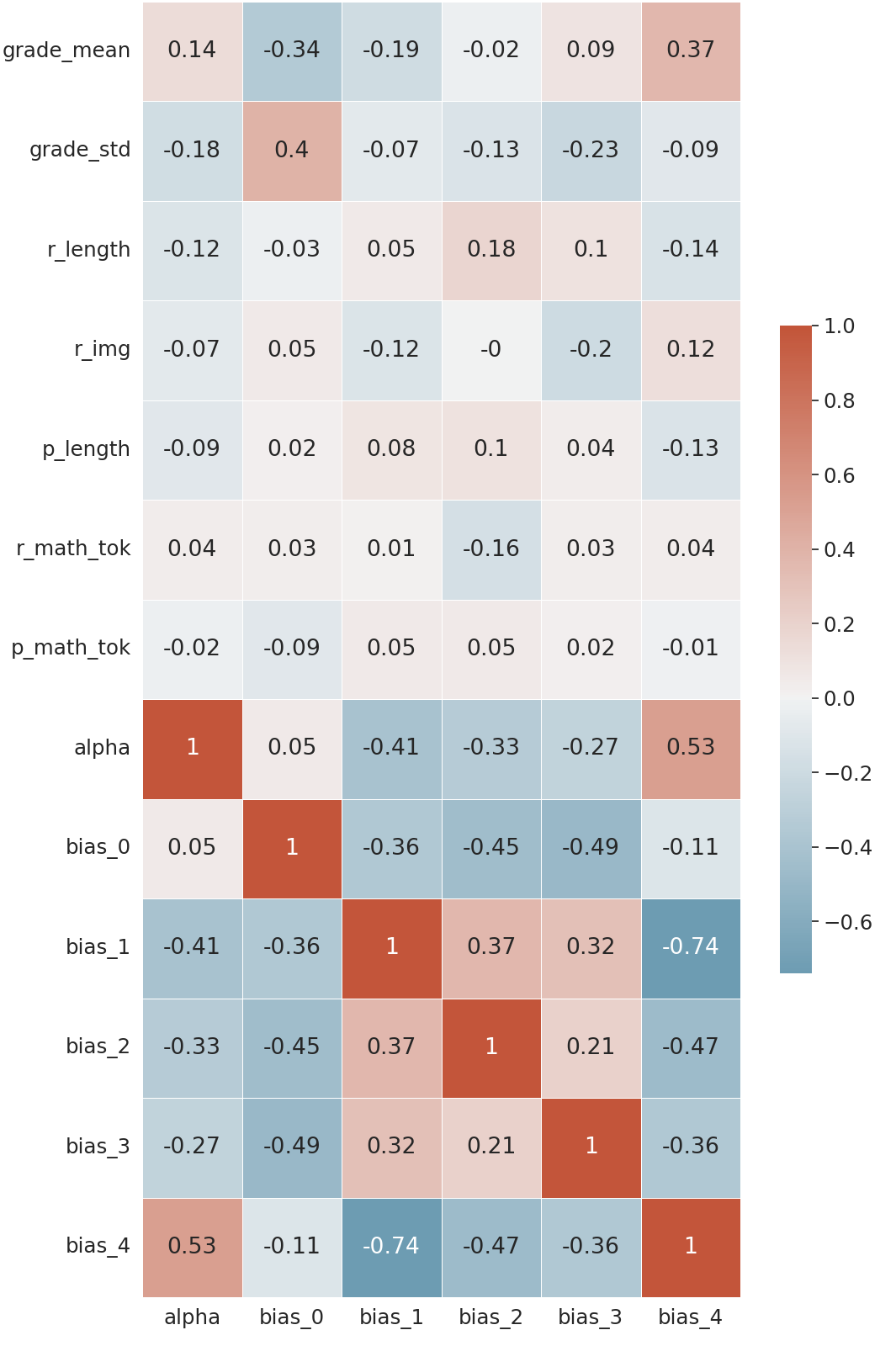}}}%
    \hspace{-.5cm}
    \subfloat[\centering P-values Matrix]{{\includegraphics[width=0.5\textwidth]{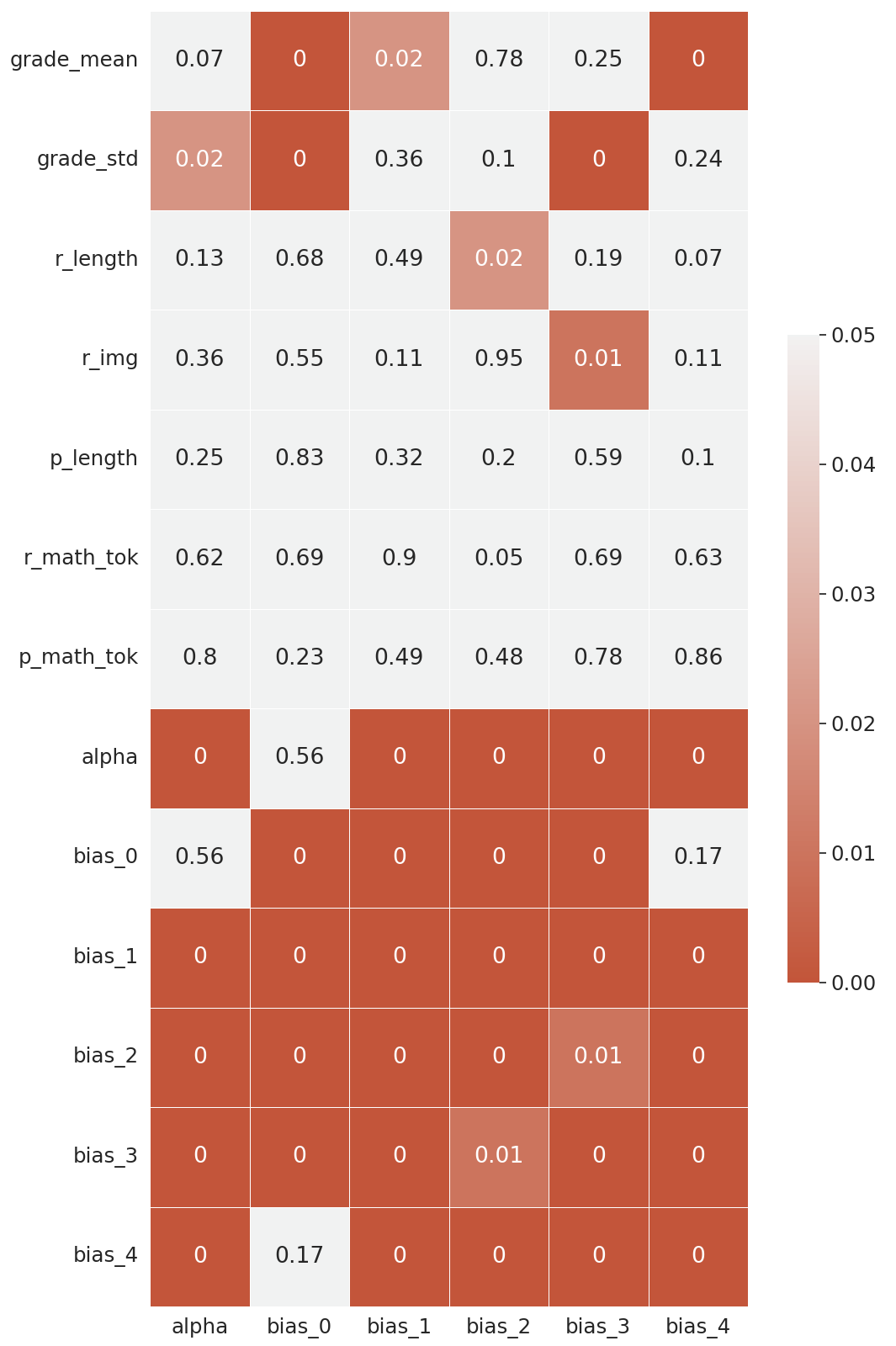}}}%
    \caption{Correlation coefficients and corresponding p-values across the bias, variance terms, and response features for the scorer-specific bias and variance models.}%
    \label{fig:corrp1}%
\end{figure*}
In Table~\ref{tab:bcj}, we see that the learned scorer biases for each cluster are highly correlated with the observed score distribution across score categories. However, it is not obvious how the variance, i.e., the inverse of the temperature parameter ($\alpha$), correlates with other model parameters and response features. Therefore, we calculate the correlation coefficient (left) and the corresponding p-value (right) between each pair of model parameters and response features and show them in Figure~\ref{fig:corrp1}. In the left part of the figure, we see that $\alpha$ positively correlates with the mean of scores and negatively correlates with the standard deviation of scores. In the right part of the figure, we see that $\alpha$ is significantly correlated with the standard deviation of scores, which is expected since this temperature parameter is designed to capture the variation in score category assignments. We also see that $\alpha$ is also significantly correlated with the bias terms of each score category, with a positive correlation with the bias for score category $4$ and negative correlation with the bias for other categories.  

For the bias terms, we see that most of the biases are significantly correlated with the mean and standard deviation of scores, but less correlated with question-response pair features. This observation suggests that the bias terms mainly depend on scorer behavior rather than the question-response pair, which is what the model intended to do; the question-response pair is captured by the $\mathbf{w}_c^T \mathbf{r}_i$ term in Eq.~\ref{eq:scalar}. The bias for score category $2$, however, does not significantly correlate with the mean and standard deviation of scores but significantly correlates with other question-response pair features. One possible explanation is that since this score category is the middle of all scores, scorers do not show any bias towards or against this score category and can solely rely on the actual content of the question and response. 
, for example, the length of the response which might show that bias 2 does not accurately represent scorer grading behavior.  

\section{Case Study: same scorer, different responses}
Table~\ref{tab:case1} shows several examples of different questions and responses and corresponding scores for a single scorer, with the actual score, biases calculated from the content-driven scorer bias and variance model, and predicted scores for different models. The overall bias for this scorer is $[-0.043, -0.36,$ $ -0.212, 0.061, 0.439]$ across all score categories, which indicates that this scorer prefers to assign high scores (especially the full score $4$) but often assigns low scores except the lowest score ($0$). Overall, we see that if we do not include biases in the AS model (the sixth column), the AS model tends to predict middling scores, while the human scorer tends to give students full credit ($4$). For Question~2, this example shows that the content-driven scorer bias model captures nuanced scorer preference: for the meaningless response ``idk'', which should have a score of $0$, the scorer has a strong preference towards giving it a high score ($3$). This bias only appears for seemingly meaningless responses but not overall (overall bias towards score category $3$ is minimal at $0.061$). Therefore, we see that the scorer-specific model cannot capture this information since its biases and variance are global across all question-response pairs for this scorer. As a result, content-driven scorer models are more flexible in handling these cases compared to other models, which is also evident in the quantitative results in Table~\ref{tab:as} that this model achieves the highest overall AS accuracy. 

\FloatBarrier
\begin{table*}[h]
\resizebox{2\columnwidth}{!}{
\begin{tabularx}{\textwidth}{|l|X|l|l|l|l|l|}
\hline
Question& Response & Actual & Content- & Scorer- & No & Content-driven scorer bias                                   \\
id & &  score&  driven & specific &  bias &  \\
 & &  &  prediction & prediction &  prediction &  \\

\hline
1    & The graph was touching the origin, but it didn't have a straight line         & 4          & 4                      & 4                       & 3               & {[}-0.61, -1.29, 0.04, -0.33, 1.33{]}  \\ \hline
2     & It meets the origin and it goes perfectly diagonal.                           & 3          & 4                      & 4                       & 3               & {[}-0.26, -1.80, -0.57, -0.39, 2.09{]} \\ \hline
            & Because it's a straight line that goes through the orgin                      & 4          & 4                      & 4                       & 3               & {[}0.13, -1.53, -0.76, -0.47, 1.71{]}  \\ \hline
            & its porportional because it has a straight line and and starts at the bottom. & 3          & 3                      & 4                       & 2               & {[}1.19, -0.20, -3.18, 1.24, 1.21{]}   \\ \hline
            & idk                                                                           & 3          & 3                      & 0                       & 0               & {[}-6.26, -0.09, 2.76, 4.56, 1.50{]}   \\ \hline
\end{tabularx}}
\vspace{.2cm}
\caption{Examples of student response and scores for a single scorer with biases $-0.043, -0.36, 0.061, -0.212, 0.439$ for all score categories. Notice that the no-bias prediction is the prediction of the content-driven model that does not scale with bias.}
\label{tab:case1}
\end{table*}
\end{document}